# Recent Technological Advances in Natural Language Processing and Artificial Intelligence

Shah,Nishal P. ; Dept of Electrical Engg, IIT Delhi

**Introduction**

There has been many philosophical discussions about the realizability of machines that are as intelligent as (or more) humans. Turing Test defines a certain methodology to test if a machine is intelligent or not.  The machine may or may not have the same processing method as humans. Though there has always been a debate regarding the validity of the test (John Searle et al), satisfying Turing test has been a goal of computer scientists from quite a long time.

A recent advance in computer technology has permitted scientists to implement and test algorithms that were known from quite some time (or not) but which were computationally expensive. Two such projects are IBM's Jeopardy as a part of its DeepQA project [1] and Wolfram's Wolframalpha[2].  Both these methods implement natural language processing (another goal of AI scientists) and try to answer questions as asked by the user. Though the goal of the two projects is similar, both of them have a different procedure at it's core.

In the following sections, the mechanism and history of IBM's Jeopardy and Wolfram alpha has been explained followed by the implications of these projects in realizing Ray Kurzweil's [3] dream of passing the Turing test by 2029. A recipe of taking the above projects to a new level is also explained.

**IBM's Jeopardy!**

 First, it is needed to understand what Jeopardy is. ***Jeopardy!*** is an American quiz show featuring trivia in history, literature, the arts, pop culture, science, sports, geography, wordplay, and more. The show has a unique answer-and-question format in which contestants are presented with clues in the form of answers, and must phrase their responses in question form [4]. Hence, to successfully understand the game one must understand the question which may be tricky using the puns and trends of human language (English in this case). After that, it must find answer. The player must also decide to buzz in or not, depending upon the confidence he has in the answer.

Now, what the IBM's system does is that it tries to understand the question (which is in the answer form) by parsing and tries to separate the components which would help it in searching for the answer. The system has large amounts of UNSTRUCTURED data in form of poems, newspaper articles, textbooks, etc spanning all the domains of the game. The system tries to find an answer through this data and comes up with many possible answers with different confidence levels. The answer with the most confidence level is selected and depending upon the confidence level of this answer, the player decides to buzz in or not. Upon answering, a sentence in form of a question is formed which involves sentence generation. [1]

Though this method works, there are some fundamental differences between how humans think and how the IBM's system works. [N]

- This method is like a brute force method just like the IBM's Deep Blue which won chess some years back. By 'brute force' method, it is meant that the method for the search of solution from the problem space (the large amount of literature and data provided to the machine) depends on the high processing speed made possible by the recent technological advances.

- The brute force method may be partially justified if the data is presented in a more 'realistic' sense. The UNSTRUCTURED data encompassing about EVERYTHING does not reflect how humans obtain, store and perceive information. There is POVERTY OF STIMULUS [5] which means that not everything is heard or read by a human in his lifetime, but still he can deduce facts and make models about language or concepts and that is how he stores the information for future use. This seems to be the case in Wolfram alpha (explained later).

- Another feature which the system has is the sequential processing. This means that the solutions are searched and narrowed down one after another. But this is clearly not the way humans find the answer. Neuroscientists believe that processing in brain is essentially parallel which means we select the probable answers in parallel and it is only in the later stage that we compare the most plausible answers and select one of them.

There is a viewpoint that it is not necessary to mimic human – form of intelligence in order to pass the Turing test. There can be other non-human intelligent designs which are functionally as good. After all, it won the Jeopardy on live television against the champions of the game.

**Wolfram Alpha:**

It is described by some reviewers as the search engine which has the potential to replace Google and start a new revolution in information exchange. Here, the user types in a query as a sentence in English seeking some facts or answers. The system then tries to understand the query and does the required operation. For example, if the request is to do a mathematical operation, it does so. [2] Like 'add 2 and 3' will give 5. At the same time, if we

ask – 'the population growth of India'. Then, it uses the data from various sources and presents it in a tabular / graphical form.

The project sounds very ambitious from the functional description given above. Let us look into the working of the system. After understanding the query of the user, and if it is doing some calculation, then the 'Mathematica' code is generated of the query which is evaluated and the answer is obtained. But for 'seeking' information, it searches for the facts in STRUCTRED data bank and retrieves the information that is relevant and organises it in a more presentable format.

It seems that the issues with using the UNSTRUCTURED and vast amount of information by the IBM Jeopardy are resolved in this case. As the semantics of the information would be used while structuring the information, it is appears to be more intelligent.

But how is this structured data acquired? Learning is what is naive about the system. The data is fed into the system using engineers and field experts consulting the data already available. In essence, it is the intelligence of the experts which is mimicked by the system. Or, the answers to every possible question are fed to the system apriori. The challenge reduces to formulating and structuring all the information available to man in its original form and storing it.

### IBM's Jeopardy versus Wolfram Alpha versus Google

Though all these systems were made up for very different purposes, it is worthwhile to compare them as all of them have one thing which is common – interacting with humans in an 'intelligent' manner.

All three of the above use large amount of computational power which reflects the fact that human brain has huge computational ability. This confirms the fact that the computational ability of the brain is the bare NECCESARY for achieving intelligence and there is no doing away with it. (Dt..) Evolution would support this fact that if there had been a more efficient (in terms of computational and energy efficiency) way to achieve this intelligence, we would have it or, we would observe the brains evolving in that direction.

So, the parameters on which we can compare these three systems and human brain is 'how' this enormous computational capability is utilised.

For explaining the difference between the three systems above, let's try to find the solution for the query – 'The capital of India.'

Google: It separates Capital and India as the keywords from the phrase by removing the commonly used phrases in English. It tries to search web-pages which have these keywords pre-assigned to them. It then ranks these pages based on the relevance which is in turn determined by the number of users that have clicked the page for same search in past.  The results are shown in the order of their ranks.

The goal of Google is to help find web-pages which can help in answering our questions. Clearly, there is no knowledge of the content of the page which is utilised by Google. Hence, the page which gives no information but is still tagged with capital and India is a plausible candidate before page ranking is applied to it.

Also, if I search – 'capital of India' and a non-sensical permutation of same – 'India of Capital' returns almost the same search results. This proves that there is not understanding of the whole phrase except that, which word is potential keyword and which is not.

IBM's Jeopardy: If we fire the same query but in Jeopardy format – 'This is the capital of India'. It understands the query that a 'concept' is to be FOUND (search in the solution space) which is related to two concepts of 'India' and being a 'Capital'. It then tries to search through all the literature it has to find the concepts which can be potentially related to the concept of India via its property of concept 'capital'.  It then calculates the confidence it has in each of the potential answers and decides what to do depending upon his strategy. Hence, it has to create a concept and identify its properties and relations to other concepts, all when the query is fired. No doubt it takes 3 minutes or so to answer the question with – 'What is New Delhi?'

Wolfram Alpha:  Here also, the same query is fired 'The Capital of India'.  It understands from the query itself that the 'capital' property of 'India' is the required solution. It then extracts the data about India's capital from it's already structured data base and presents the information of Delhi in a structured form to the user.

In essence, Google is a tool for FINDING the solution, Wolfram alpha is a tool for ANSWERING [2] the solution and IBM's Jeopardy is a tool for ANSWERING BY FINDING.

**On the future..**

The above techniques are very good from a functional point of view. They already have the potential to be implemented in various domains like Medical Expert systems, etc.

But they lack the ingenuity of a human brain. For overcoming this, one can try to implement the models of cognition and brain put forward by Neuroscientists and Cognitive Scientists.

Another solution is to try to implement a Child Turing Machine and trace the evolution of the brain of child from a nearly blank slate (with some characters already written – like innate features) to an alert and interacting brain.

Hence, it must create a model for the data it comes across during its lifetime and model it using a new one or fitting it in an already existing one. This model shall be the sole means by which the information is stored. This is the essence of bootstrapping. That is, it decides on its own what model to form and how to form it. This is different from Wolfram alpha where the model is already formed or IBM's Jeopardy where no permanent model is ever formed. Again, relying on cognitive scientists has serious philosophical risks.

Another issue to be resolved is the quality of the data provided to the system.

What if the data is collected by different experts who have bias, or by two experts who have opposite opinions? [2]

We can resolve this by feeding into the system all the possible alternates. But to make it more human like in it's approach, it needs to identify those biases on it's own and store alternates without the aid from human expert.

The system can handle quantitative facts or qualities better, but what about the qualitative ones?[2]

We can use fuzzy logic. But this seems to handle qualities quantitatively. This again leads to a deadlock.

These problems signify the importance of Natural Language Processing which not only helps in understanding the queries typed in by the user, but also helps in the RAW data available to it to make models both online and offline.

Hence, a balanced mixture of both IBM's Jeopardy and Wolfram Alpha will be a significant improvement for interaction with computers for exchange of information and otherwise.